\documentclass{article}

\usepackage[preprint]{neurips_2026}

\usepackage[utf8]{inputenc} 
\usepackage[T1]{fontenc}    
\usepackage{url}            
\usepackage{booktabs}       
\usepackage{amsfonts}       
\usepackage{nicefrac}       
\usepackage{microtype}      
\usepackage{xcolor}         

\usepackage{xspace}
\usepackage{natbib}
\usepackage{epigraph}
\usepackage[most]{tcolorbox}
\usepackage{subcaption}
\usepackage{tabularx}
\usepackage{makecell}
\usepackage{fontawesome5}
\usepackage{algorithm}
\usepackage{algorithmic}
\usepackage{wrapfig}
\definecolor{mycitecolor}{HTML}{3498DC}
\definecolor{mylinkcolor}{HTML}{E74D3B}
\definecolor{myblue}{HTML}{4594c1}
\definecolor{myurlcolor}{HTML}{980000}
\usepackage[colorlinks=true,linkcolor=mylinkcolor,citecolor=mycitecolor,urlcolor=myurlcolor]{hyperref}

\definecolor{darkblue}{rgb}{0.0, 0.17, 0.58}
\usepackage{titletoc}
\usepackage{array}
\usepackage{colortbl}
\newtcolorbox{motivationbox}[1]{
    colback=gray!2!white,
    colframe=black!38!gray,
    arc=2mm, auto outer arc,
    fonttitle=\bfseries,
    title=#1,
    boxrule=0.8pt,
    left=5pt,
    right=5pt,
    top=5pt,
    bottom=5pt
}
\usepackage{enumitem}
\definecolor{c1}{HTML}{4a1486}
\definecolor{c2}{HTML}{99000d}
\definecolor{c3}{HTML}{005a32}
\definecolor{c4}{HTML}{084594}

\usepackage[capitalize,noabbrev]{cleveref}

\title{STRIDE: Learnable Stepwise Language Feedback \\ for LLM Reasoning}

\author{
\textbf{Junjie Zhang\textsuperscript{1}\thanks{Email: \texttt{junjie.zhang@ntu.edu.sg}}}\;,
\textbf{Guozheng Ma\textsuperscript{1}}, 
\textbf{Shunyu Liu\textsuperscript{1}}, 
\textbf{Zetian Hu\textsuperscript{1}},
\textbf{Yongcheng Jing\textsuperscript{1}}, \\
~\textbf{Ting-En Lin\textsuperscript{2}},
\textbf{Yongbin Li\textsuperscript{2}\thanks{Corresponding authors: \texttt{shuide.lyb@alibaba-inc.com}, \texttt{dacheng.tao@ntu.edu.sg}}}\;,
\textbf{Dacheng Tao\textsuperscript{1}\footnotemark[2]}\\[6pt]
\textsuperscript{1}Generative AI Lab, College of Computing and Data Science, Nanyang Technological \\~~University, Singapore 639798 \\
\textsuperscript{2}Tongyi Lab, Alibaba Group
}

\begin{document}

\newcommand{\method}{STRIDE\xspace}
\newcommand{\methodtexttt}{\texttt{STRIDE}\xspace}
\newcommand{\methodbf}{\textbf{STRIDE}\xspace}

\newcommand{\pz}{\phantom{0}}
\newcommand{\grayrow}{\rowcolor{gray!10}}
\newcommand{\cmark}{\textcolor{c3}{\faCheck}}
\newcommand{\xmark}{\textcolor{c2}{\faTimes}}

\maketitle

\begin{abstract}

Recent advances in Reinforcement Learning~(RL) have underscored its potential for incentivizing reasoning capabilities of Large Language Models~(LLMs). 
However, existing step-level efforts suffer from costly annotations that limit domain coverage, while scalar scores further impose an information bottleneck, offering insufficient semantic bandwidth to improve intermediate decisions. Alternative language-critique approaches, which rely on frozen or external critics, provide richer textual feedback but lack the scalability needed for sustained policy improvement.
In this work, we propose language-driven stepwise trajectory redirection, termed as \methodbf, a novel training framework that shifts process supervision from scalar rewards to \textit{learnable} stepwise language feedback. 
Specifically, we co-train a generator and a generative verifier using only outcome-based rewards, eliminating external annotations, while delivering sustained policy improvement through jointly aligned verifier training. 
The verifier's stepwise language critiques explicitly localize and explain failures, enabling the generator to redirect reasoning trajectories at intermediate steps toward alternative decisions. 
The trajectory redirection design guarantees harmless policy improvement, even under noisy or suboptimal verifier feedback. 
Experiments on diverse reasoning benchmarks show that \method significantly outperforms state-of-the-art baselines, as well as achieving breakthroughs on zero-pass-rate problems where scalar methods yield no learning signal in our ablation studies, demonstrating the effectiveness of learnable stepwise language feedback for enhancing LLM reasoning.

\end{abstract}

\section{Introduction}

The recent surge in reasoning capabilities of LLMs has been largely driven by Reinforcement Learning from Verifiable Rewards~(RLVR)~\cite{Guo_2025, shao2024deepseekmath_grpo, ouyang2022training_rlhf}. However, these methods rely on sparse, outcome-based rewards~\cite{ouyang2022training_rlhf, cobbe2021trainingverifiers_gsm8k}, which offer no feedback on individual reasoning steps, leaving credit assignment a fundamental unsolved challenge in multi-step reasoning.

Process Reward Models~(PRMs)~\cite{lightman2023let, solvingmathwordproblems, cui2025process} advance credit assignment through step-level supervision, yet suffer from two compounding limitations. 
First, reliable step-level annotations are prohibitively 
expensive to obtain~\cite{lightman2023let}, confining PRMs to narrow domains. Automated labeling alleviates the cost but introduces inaccurate labels that actively mislead training with harmful gradient noise~\cite{setlur2024rewarding, cui2025process}. 
Second, scalar scores impose a fundamental representational constraint: compressing high-dimensional reasoning into a single numerical value creates an information bottleneck~(see~\S\ref{sec:ib_bottleneck}), providing insufficient semantic bandwidth to distinguish or correct qualitatively different error modes~\cite{solvingmathwordproblems}.
To overcome the representational constraint, critique-based methods~\cite{zhang2025critgrpo, selfcorrect} and SFT-based error-corrective approaches~\cite{xi2024automathcritique, pan2025lemma, backtracking} shift supervision from scalar to language, recovering the semantic richness that scalars discard. 
However, their reliance on frozen or external critics and on supervised fine-tuning limits adaptability, preventing sustained improvement as the policy evolves. 
Most recently, TANGO~\cite{zha2025rltango} co-trains a generative verifier alongside the generator, yet converts its language output back into step-level scalar rewards, reintroducing the same information bottleneck. 
A natural question then arises: can a feedback mechanism that is simultaneously stepwise, language-informative, and learnable resolve all of the above limitations?

In this paper, we propose \methodbf to answer this question. \method co-trains a generator and a generative verifier using only outcome-based rewards, requiring no step-level annotations. The core insight is the shift of process supervision paradigm from scalar reward signals to \textit{learnable} in-context language feedback: language critiques from the co-trained verifier carry the semantic direction needed to localize and rectify specific reasoning errors, and generate productive training signal even on hard problems where scalar methods yield identically zero advantage~\cite{shao2024deepseekmath_grpo, yu2025dapo}. Specifically, the framework operates through an interleaved three-phase schedule: Base Policy Optimization (Phase~I), Generative Verifier Optimization (Phase~II), and Guided Trajectory Redirection (Phase~III). For challenging problems where the generator fails, \method localizes the First Point of Failure~(FPF) and employs a \textbf{Multi-Point Redirection Strategy}, efficiently constraining the search space by redirecting from verified prefix steps. To ensure training stability, \method maintains \textbf{outcome-only reward grounding}, where learning signals are strictly tied to final correctness, shielding the model from the harmful gradient noise that unreliable step-level signals introduce~\cite{zhang2025critgrpo}. By unlocking the information bandwidth of process supervision, \method enables LLMs to overcome reasoning plateaus through guided self-correction rather than exhaustive sampling.

\begin{figure}[!t]
    \centering
    \includegraphics[width=\linewidth]{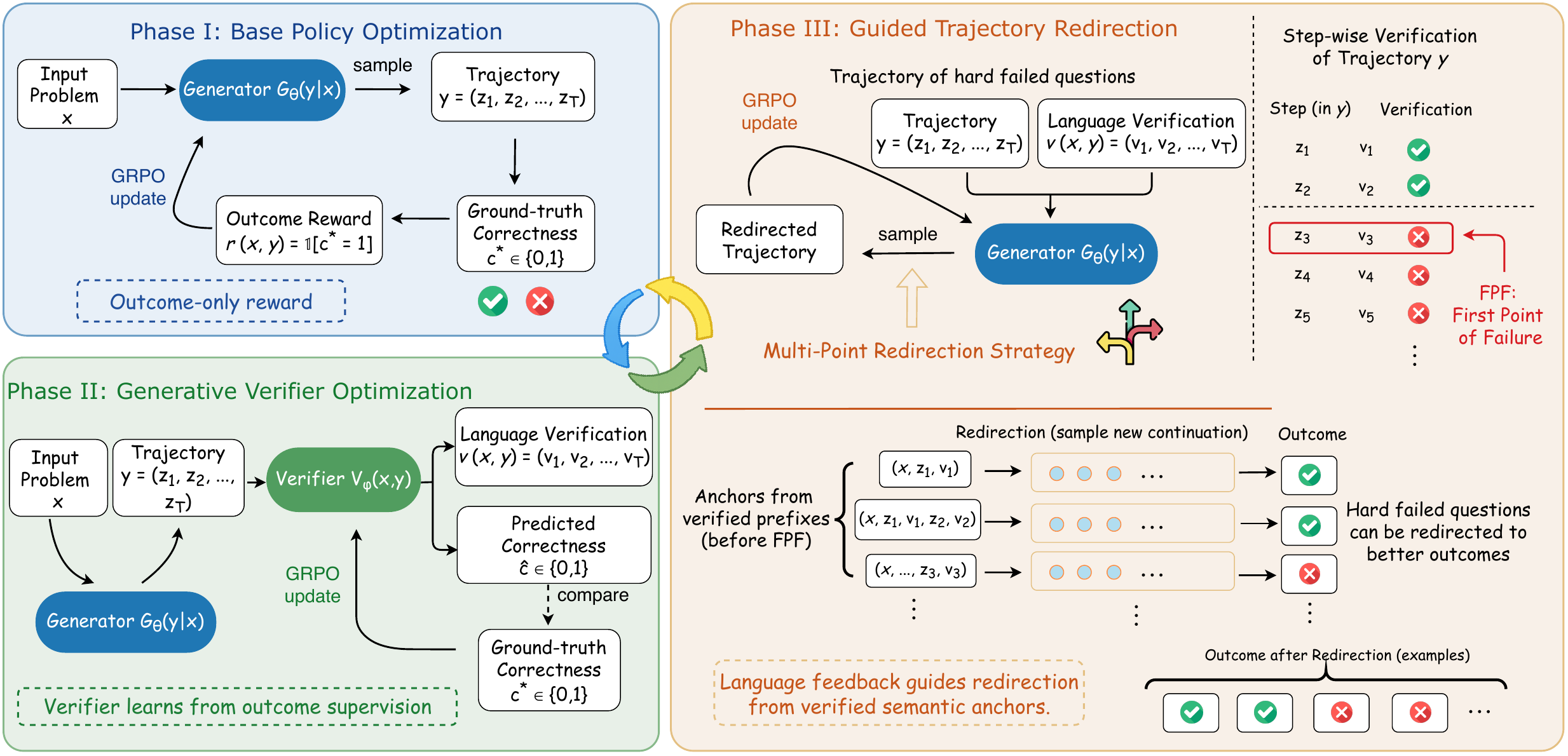}
    \caption{\textbf{Overview of the \methodbf framework.} 
    \methodbf shifts the process supervision paradigm from unidimensional scalar rewards to high-bandwidth in-context guidance. 
    \textbf{Phase I} builds basic reasoning capabilities through outcome-based GRPO. 
    \textbf{Phase II} optimizes a generative verifier to decompose terminal rewards into step-level linguistic feedback $v_t$. 
    \textbf{Phase III} leverages the verifier to localize the First Point of Failure (FPF) and triggers Multi-Point Redirection. 
    By initiating reconstruction from multiple verified anchors with actionable semantic guidance, the framework effectively constrains the vast exploration space to break through reasoning stagnation.}
    \label{fig:stride_framework}
\end{figure}

The main contributions of this work are as follows:

\begin{itemize}[leftmargin=*]
    \item \textbf{Paradigm Shift to High-Bandwidth Feedback:} We first propose shifting process supervision toward high-bandwidth with learnable stepwise language feedback to unlock richer training signals.
    
    \item \textbf{The \method Framework:} We introduce an interleaved three-phase co-training schedule that incorporates a \textit{Multi-Point Redirection} strategy. This approach utilizes verified semantic anchors to efficiently constrain the reasoning exploration space at verified prefix steps.
    
    \item \textbf{Superior Empirical Performance:} We demonstrate that \method significantly outperforms reward-based baselines across diverse reasoning benchmarks, achieving consistent gains at minimal additional training cost: Phase~III accounts for only 1/13 of the total training schedule yet delivers the decisive capability to learn from previously unsolvable problems.
\end{itemize}

\section{Related Work}

\begin{table}[t]
\centering
\small
\caption{Comparison of \method and closely related method categories across four key design dimensions.
Policy update: whether the method updates the base policy (vs.\ inference-only).
Language feedback: whether language-form feedback is directly utilized in the training loop.
Learnable verifier: whether a dedicated verifier is jointly optimized with the generator via RL.
RL-based training: whether RL is used for policy optimization.}
\label{tab:comparison}
\setlength{\tabcolsep}{4pt}
\begin{tabular}{lcccc}
\toprule
\textbf{Method} & \makecell{\textbf{Policy}\\\textbf{update}} & \makecell{\textbf{Language}\\\textbf{feedback}} & \makecell{\textbf{Learnable}\\\textbf{verifier}} & \makecell{\textbf{RL-based}\\\textbf{training}} \\
\midrule
Outcome-based RL~\citep{shao2024deepseekmath_grpo,xue2025comas}          & \cmark & \xmark                                       & \xmark & \cmark \\
Inference-time Search~\citep{tot,backtracking}                            & \xmark & \xmark                                       & \xmark & N/A    \\
Inference-time Verifier Search~\citep{khalifa2025process,setlur2024rewarding} & \xmark & \cmark\ {\footnotesize(inference)}           & \xmark & N/A    \\
RL w/ Language Scaffold~\citep{shi2026r}                                  & \cmark & \xmark\ {\footnotesize(distilled)}           & \xmark & \cmark \\
Critique-based RL~\citep{zhang2025critgrpo,selfcorrect,kumar2025score}     & \cmark & \cmark\ {\footnotesize(frozen/self)}         & \xmark & \cmark \\
SFT Error-Corrective~\citep{pan2025lemma,xi2024automathcritique}          & \cmark & \cmark\ {\footnotesize(SFT)}                 & \xmark & \xmark\ {\footnotesize(SFT)} \\
Co-training w/ Scalar~\citep{zha2025rltango}                              & \cmark & \xmark\ {\footnotesize(scalar)}              & \cmark & \cmark \\
\midrule
\grayrow \methodbf                                                        & \cmark & \cmark\ {\footnotesize(co-trained)}          & \cmark & \cmark \\
\bottomrule
\end{tabular}
\end{table}

\textbf{RLVR and Process Supervision.}
RLVR has demonstrated significant efficacy in enhancing the reasoning performance of LLMs. Early paradigms primarily rely on Outcome-based Reward Models (ORMs) \cite{ouyang2022training_rlhf}, where the model is optimized using a terminal reward signal derived from ground-truth correctness \cite{Guo_2025, cobbe2021trainingverifiers_gsm8k}. While ORMs provide an unbiased supervision signal, they suffer from severe credit assignment challenges, as the model receives no feedback on which specific steps in a multi-step trajectory led to the final success or failure.
To mitigate this, recent research shifted toward Process Supervision, introducing Process Reward Models (PRMs) that assign scalar scores to individual reasoning steps \cite{lightman2023let, chen2024step, cui2025process, solvingmathwordproblems, mathshepher, zeng2025versaprm}.
However, despite their density, these PRMs remain confined to a unidimensional scalar space. As discussed in \Cref{sec:ib_bottleneck}, compressing high-dimensional logical reasoning into a single numerical value creates an information bottleneck, leading to representational collapse where distinct error modes become indistinguishable \cite{solvingmathwordproblems}. Consequently, while PRMs improve credit assignment, they lack the semantic bandwidth necessary to guide the model through complex logical redirections, a gap that  \methodbf  aims to fill.

\textbf{Generative Verification and Language Feedback.}
In the context of LLMs, a growing body of work explores the use of generative verification and language feedback. Unlike discriminative models, generative verifiers provide feedback in the form of natural language critiques, which offer higher informational density \cite{zhang2025simple,critiqueoutloud}. This paradigm shift is motivated by the observation that LLMs often possess latent knowledge of their errors that cannot be fully expressed through a single numerical score \cite{huang2024selfcorrection}.
Existing approaches generally fall into two categories. First, Inference-time Refinement methods~\cite{reflexion,selfeditor},
leverage linguistic feedback to iteratively correct reasoning paths during the decoding stage. However, these methods primarily focus on improving a single instance at test time rather than updating the underlying policy. Second, Alignment via Feedback methods~\cite{rlaif,jiang2025pagmultiturnreinforcedllm,selfcorrect,kumar2025score,xie2025ctrl,liu2025trustverify},
attempt to internalize feedback during the training phase. However, a major challenge in this area is the instability of linguistic signals: without a robust grounding mechanism, generative feedback can lead to \textit{hallucinated gradients} where the model optimizes toward incorrect critiques \cite{zhang2025critgrpo}. \methodbf distinguishes itself by integrating generative verification directly into an interleaved RL training loop and employing an outcome-only reward, ensuring that high-bandwidth language feedback leads to stable and verifiable policy improvements.

\textbf{Refining, Rethinking and Trajectory Redirection.}
The concept of refining or rethinking a trajectory after an initial attempt is a well-established strategy for solving complex reasoning tasks. Conventional methods typically employ Inference-time Search~\cite{zhang2025supervised,snell2025scaling}, such as Tree-of-Thought (ToT)~\cite{tot}, Backtracking Search~\cite{backtracking}, and SWE-Search~\cite{antoniades2024swe}, which explore multiple reasoning branches or solution paths to identify valid outcomes. While effective during decoding, these approaches do not inherently improve the model's base policy. In the training context, methods like STaR \citep{zelikman2022star}, Quiet-STaR \citep{quietstar}, and DOTS \citep{yue2025dots} focus on self-taught reasoning by fine-tuning on successful \textit{rationale} trajectories. However, these frameworks often ignore the valuable signal present in failed attempts, treating them as simple negative samples rather than opportunities for learning error correction.
\method draws inspiration from this lineage but introduces a fundamental shift through Guided Trajectory Redirection. 
Unlike Re-sampling, Re-Reading~\cite{rereading}, or Re-solving~\cite{wang2026re2} strategies that often restart the reasoning process from scratch, our Multi-Point Redirection leverages the verifier's localization of the First Point of Failure (FPF) to pinpoint where the logic diverged. This allows the model to re-explore only the necessary sub-trees of the reasoning space, significantly constraining the exploration effort compared to unguided trial-and-error \cite{huang2024selfcorrection}. Furthermore, by providing explicit in-context language feedback at the redirection anchors, \methodbf transforms the \textit{refine} process from a stochastic search into a directed evolution of the reasoning policy.

\textbf{Positioning \method in the Landscape.}
\Cref{tab:comparison} provides a structured comparison of \methodbf against closely related methods across four dimensions. \method is the only approach that simultaneously satisfies all four properties, unified by a single design principle: shifting process supervision from scalar rewards to in-context language feedback produced by a co-trained verifier.

\section{Preliminaries}
\subsection{The Generator-Verifier Framework}
\label{sec:prelim_gv}

The Generator-Verifier (GV) framework~\citep{zha2025rltango} uses RL to concurrently train a generative \textit{Generator} $G_\theta$ and a generative \textit{Verifier} $V_\phi$. Given a query $x$, the generator produces a reasoning path $y = (z_1, \dots, z_T)$ of sequential thought steps; the verifier assesses each step and produces a language verification sequence $v = (v_1, \dots, v_T) = V_\phi(x, y)$, from which step-level correctness labels are parsed. Both models are trained via RLVR using only the outcome signal (whether the final judgment $\hat{c}_O$ matches the ground truth $c_O^*$), with no access to intermediate step annotations.

To update the generator, prior work combines step-level and outcome-based advantages via a decaying coefficient $\alpha$. However, this design carries a critical instability: step-level rewards derived from the verifier are difficult to align with the ground-truth outcome signal~\citep{zha2025rltango}, leading to unreliable gradient updates. Formal definitions and the full optimization objective are provided in Appendix~\ref{app:prelim_formal}.

\subsection{The Information Bottleneck of Scalar Rewards}
\label{sec:ib_bottleneck}

In the GV framework, the verifier compresses a high-dimensional reasoning path $y = (z_1, \dots, z_T) \in \mathcal{Y}$ into a unidimensional scalar reward sequence $r \in \mathbb{R}$. Since $\text{dim}(\mathcal{Y}) \gg \text{dim}(\mathbb{R})$, this mapping is inherently \textbf{many-to-one}: paths with fundamentally different logical errors may receive identical scalar values, providing the generator no semantic direction to identify \textit{where} or \textit{why} a mistake occurred. A formal Rate-Distortion analysis showing that scalar rewards are fundamentally limited in information bandwidth is provided in Appendix~\ref{app:ib_formal}.

The fundamental limitation of scalar rewards lies in this \textit{information-theoretic disparity} between high-dimensional reasoning and unidimensional rewards, rendering the error-correction process ill-posed. As illustrated in Appendix~\ref{app:rep_collapse}, two semantically distinct errors collapse to the same scalar value, while language feedback restores the missing semantic dimension. To handle this challenge, we introduce the \method framework in the following section.

\section{Methodology}
\label{sec:method}
In this section, we present \methodbf, a novel three-phase training framework that effectively trains, generates, and leverages verification for stepwise redirection, overcoming the performance plateau of LLM with purely scalar reward training. The core idea of \method is to shift the process supervision paradigm from scalar rewards to stepwise language feedback, enabling the generator to improve with informative stepwise language feedback during the training process.

\subsection{Overview of \method Framework}
The \method framework is a unified, three-phase training system designed to evolve the reasoning capabilities of LLMs from simple outcome matching to active, guided redirection. Unlike traditional co-training paradigms, \method executes these phases in an interleaved manner with a scheduled cadence (e.g., a $9:3:1$ ratio for $G$ training, $V$ training, and $G$ redirection), ensuring a stable progression from base policy optimization to complex error rectification. 

As depicted in \Cref{fig:stride_framework}, the framework orchestrates the interaction between the Generator $G_\theta$ and the Verifier $V_\phi$ through the following stages:

\textbf{Phase I: Base Policy Optimization.} In this foundational stage, the generator $G_\theta$ is trained using Group Relative Policy Optimization (GRPO) based on outcome rewards $c_O^*$. This phase occupies the largest portion of the training cycle (9/13 of the schedule), focusing on building the fundamental ability of the model to generate coherent reasoning trajectories $y$ that reach the correct final answer.

\textbf{Phase II: Generative Verifier Optimization.} Following the base generator updates, the verifier $V_\phi$ is trained in an outcome-based RLVR manner to provide generative verification $v = (v_1, \dots, v_T)$. By approximating the outcome-based supervision of overall correctness prediction (whether $\hat{c}_O = c_O^*$), the verifier learns to decompose the terminal signal into stepwise language verification $v_t \in \mathcal{V}^*$. 

\textbf{Phase III: Guided Trajectory Redirection.} This stage drives Generator $G_\theta$ to overcome reasoning stagnation by leveraging $V_\phi$ as a contextual navigator. For samples where $G_\theta$ fails and $V_\phi$ correctly identifies the error, we construct a redirection pipeline. In this phase, the generator is trained specifically to rectify its reasoning path $y$ before the first point of failure $t^*$, using the verifier's guidance $v_{t^*}$ as an in-context trigger. Crucially, this phase uses a pure redirection distribution without mixing Phase I samples to maximize the gradient focus on error correction and avoid the off-policy issues~\cite{yan2025offpolicyguidance, zhang2025critgrpo}.

\subsection{Generative Verification and Stepwise Error Localization}
This section formalizes how the verifier $V_\phi$ transitions from a passive reward model to an active error-localization tool.

\textbf{Structured Verification Generation.} 
For each step $z_t$ in a trajectory $y$, the verifier $V_\phi$ decodes a sequence of language verification $v_t$. This generative process is modeled as:$$(v_1, v_2, \dots, v_T) = V_\phi(x, y)$$The resulting sequence $v = (v_1, \dots, v_T)$ provides a high-fidelity audit trail of the generator's reasoning process. 

\textbf{The Triggering Function for Redirection.}
To automate the redirection process, we define a Triggering Function $\tau(v_t)$ that parses the semantic content of each verification step:$$\tau(v_t) = 
\begin{cases} 
0, & \text{if } v_t \text{ identifies a logical or arithmetic fallacy} \\
1, & \text{otherwise}
\end{cases}$$

The system then identifies the First Point of Failure (FPF), denoted as $t^*$:$$t^* = \min \{ t \mid \tau(v_t) = 0 \}$$This $t^*$ serves as the temporal anchor for redirection, ensuring that the generator's redirection starts at where the logic deviated, enabling precise and contextually relevant corrections.

\subsection{Guided Trajectory Redirection}
In Phase III, \method transforms flawed reasoning paths into high-value training signals by Guided Trajectory Redirection. Instead of treating the First Point of Failure (FPF) as a terminal error, we leverage it as a sign for parallel path reconstruction.

\textbf{Multi-Point Redirection Strategy.} 
Given an initial reasoning trajectory $y = (z_1, \dots, z_T)$ where the verifier $V_\phi$ localizes the first point of failure at index $t^*$, we do not merely rectify the specific step $z_{t^*}$. Instead, we define a set of anchor points encompassing the entire prefix up to the failure: $\mathcal{A} = \{ t \mid 1 \le t \le t^* \}$. For each $(x,y)$, we simultaneously construct $t^*$ distinct redirection samples. This dense sampling strategy addresses three critical challenges in reasoning alignment: (i) Deep Error Attribution: It accounts for \textit{latent drift}, where the terminal fallacy at $t^*$ is a downstream consequence of a suboptimal (though not yet incorrect) choice at $t < t^*$. (ii) Verification Noise Tolerance: It mitigates the inherent uncertainty of $V_\phi$. By re-sampling from steps preceding the detected error, the system remains robust even if the verifier fails to pinpoint the exact step of deviation. (iii) Exploration Density: It encourages the generator to explore alternative valid reasoning paths, effectively balancing error correction with path diversification.

\textbf{Context Construction.}
For each anchor $t \in \mathcal{A}$, we construct a unique redirection context $S_{redirect}^{(t)}$. The semantics of the guidance are conditioned on the anchor's position relative to the failure point:\begin{equation}S_{redirect}^{(t)} = (x, z_{1}, v_{1}, \dots, z_{t-1}, v_{t-1}, \text{Instr}^{(t)})\end{equation}where the redirection instruction $\text{Instr}^{(t)}$ is defined with subtle but crucial differences: (i) Rectification Prompt (if $t = t^*$): The verifier provides $v_{t^*}$ identifying the specific error. The generator is prompted to rectify the fallacy and resume reasoning. (ii) Exploration Prompt (if $t < t^*$): Since step $z_t$ was deemed correct but still led to a failed outcome, the generator is prompted to continue the reasoning from this valid prefix step, implicitly encouraging the discovery of more robust or efficient paths. Detailed prompt templates are provided in Appendix~\ref{app:prompts}.

\textbf{Training on Pure Redirection Distributions.}
In Phase III, we update $G_\theta$ solely on redirected samples $\{y_{redirect}^k\}_{k=1}^K$ by GRPO with rollout batch size $K$. By maintaining a non-mixed distribution from Phase I, we specialize the policy in \textit{listening} to contextual guidance for redirection. Following our robust training principle, the advantage $\hat{A}_k$ is calculated strictly based on the outcome correctness $c^*$ of each redirected trajectory: \begin{equation} \hat{A}_k = \frac{c_k^* - \text{mean}(c_1^*, \dots, c_K^*)}{\text{std}(c_1^*, \dots, c_K^*) + \epsilon} \end{equation}

If the verifier's guidance is hallucinated or logically wrong, the resulting batch trajectories $\{y_{redirect}^k\}_{k=1}^K$ are incorrect still, all yielding $\hat{A}_k = 0$ for these samples. This ensures that while we push the reasoning ceiling via correct guidance, we do not pollute the policy with verifier-induced noise, which is unavoidable if directly using scalar rewards from $V_\phi$ for advantage computation~\cite{zha2025rltango}.

\textbf{In summary,} \method establishes an interleaved three-phase training paradigm, as illustrated in \Cref{alg:interleaved}, to harmonize reasoning and verification. The core idea lies in shifting the process supervision paradigm from unidimensional scalar rewards to high-bandwidth stepwise language feedback, effectively breaking the information bottleneck in complex reasoning tasks. Benefiting from the outcome-only reward, \method maintains high robustness against verifier hallucinations by ensuring only successful redirections contribute to policy updates. Although Phase III represents a small fraction of the training cycle, it serves as the decisive engine for transcending reasoning ceilings by fostering sparse but vital \textit{breakthrough} samples that enable the model to overcome the performance plateau, as demonstrated in our experiments.

\section{Experiments}
\textbf{Models and Baselines.}
To evaluate the efficacy of \method, we employ two series of generator-verifier pairs: (1) Qwen, using Qwen2.5-Math-7B~\cite{yang2024qwen25mathtechnicalreportmathematical} as the generator and Qwen2.5-7B~\cite{yang2024qwen25} as the verifier; (2) Llama, using Llama-3.1-8B~\cite{grattafiori2024llama} for both roles. We compare our method against two primary classes of baselines: (1) Outcome-based RL: A standard RLVR approach using only terminal ground-truth rewards via GRPO~\cite{shao2024deepseekmath_grpo}. (2) TANGO~\cite{zha2025rltango}: The state-of-the-art co-training framework that utilizes the verifier to provide scalar step-level rewards alongside outcome rewards. This setup allows us to directly measure the gain from shifting from scalar-based process supervision to our proposed stepwise language feedback.

\textbf{Datasets and Benchmarks.} We conduct evaluation on five competition-level mathematical benchmarks: AIME 2024/2025~\cite{aime2024,aime2025}, AMC 2023~\cite{amc2023}, MATH-500~\cite{lightman2023let}, and OlympiadBench~\cite{he2024olympiadbench}. To assess general reasoning and robustness across domains, we further include BoardgameQA (logic)~\cite{kazemi2023boardgameqa}, CRUXEval (code)~\cite{gu2024cruxeval}, StrategyQA (commonsense)~\cite{geva2021did}, and TableBench (tabular reasoning)~\cite{wu2025tablebench}. These benchmarks represent a comprehensive testbed for complex, multi-step logical deduction.

\begin{table*}[!t]
\setlength{\tabcolsep}{4pt}
\renewcommand{\arraystretch}{1.1}
\caption{\small \textbf{Comprehensive performance comparison with prior methods} on mathematical and general reasoning benchmarks. \methodbf achieves state-of-the-art performance among 7B/8B-scale models across both domains. For mathematical reasoning, results for most baseline models are sourced from their respective original papers or the prior works \cite{guan2025rstar, shen2025satori}. We adopt the performance of reproduction of PRIME \cite{cui2025process} reported in~\cite{zha2025rltango}.
}
\vspace{0cm}
\label{tab:main}
\small
\begin{center}
\resizebox{\linewidth}{!}{
\begin{tabular}{lccccccccccc}
\toprule[1pt]
& \multicolumn{6}{c}{\textbf{Mathematical Reasoning}} & \multicolumn{5}{c}{\textbf{Out-of-Domain Reasoning}}  \\
\cmidrule(lr){2-7} \cmidrule(lr){8-12} \\[-1.0em]
\textbf{Model} & \makecell{\textbf{MATH}\\\textbf{500}} & \makecell{\textbf{AIME}\\\textbf{2024}} & \makecell{\textbf{AIME}\\\textbf{2025}} & \makecell{\textbf{AMC}\\\textbf{2023}}  & \makecell{\textbf{Olympiad}\\\textbf{Bench}} & \textbf{Avg.} & \textbf{BGQA} & \makecell{\textbf{CRUX}\\\textbf{Eval}} & \makecell{\textbf{Strategy}\\\textbf{QA}} & \makecell{\textbf{Table}\\\textbf{Bench}} & \textbf{Avg.} \\ \midrule\midrule
\multicolumn{12}{l}{\textit{Frontier LLMs}\vspace{0.02in}} \\
\pz\pz GPT-4o~\cite{hurst2024gpt} &  76.6 & 9.3 & - & 47.5 & 43.3 & - & - & - & - & -& - \\
\pz\pz Claude3.5-Sonnet~\cite{anthropic2024claude35} & 78.3 & 16.0 & - & - & - & - & - & - & - & -& - \\
\pz\pz o1-preview~\cite{jaech2024openai_o1} & 85.5 & 44.6 & - & 90.0 & - & - & - & - & - & -& -\\
\pz\pz o1-mini~\cite{jaech2024openai_o1} & 90.0 & 56.7 & - & 95.0 & 65.3 & - & - & - & - & -& - \\
\arrayrulecolor{gray}\cmidrule(lr){1-12}
\multicolumn{12}{l}{\textit{Open-sourced reasoning LLMs (large)}\vspace{0.02in}} \\
\pz\pz Llama-3.1-70B-Instruct~\cite{grattafiori2024llama} & 68.0 & 13.3 & - & 42.5 & 29.4 & - & 58.3 & 59.6 & 88.8 & 34.2 & - \\
\pz\pz OpenMath2-Llama3.1-70B~\cite{toshniwal2024openmath2} &  71.8 &  13.3 & - & 45.0 & 30.1 & - & 68.7 & 35.1 & 95.6 & 46.8 & - \\
\pz\pz NuminaMath-72B-CoT~\cite{numina_math_72b} & 64.0 & 3.3 & - & 70.0 & 32.6 & - & - & - & - & - & - \\
\pz\pz Qwen2.5-Math-72B-Instruct~\cite{yang2024qwen25mathtechnicalreportmathematical} & 82.6 & 23.3 & - & 70.0 & 49.0 & - & - & - & - & - & - \\
\pz\pz QwQ-32B-Preview~\cite{qwq-32b-preview} & 90.6 & 50.0 & 33.3 & 77.5 & 61.2 & 62.5 & 71.1 & 65.2 & 88.2 & 51.5 & 69.0 \\
\arrayrulecolor{gray}\cmidrule(lr){1-12}
\multicolumn{12}{l}{\textit{Open-sourced reasoning LLMs (small)}\vspace{0.02in}} \\
\pz\pz Llama-3.1-8B-Instruct~\cite{grattafiori2024llama} &  51.9 &  3.3 & 3.3 & 22.5 & 15.1 & 19.2 & 50.3 & 38.5 & \textbf{92.2} & 32.4 & 53.4 \\
\pz\pz OpenMath2-Llama3.1-8B~\cite{toshniwal2024openmath2} & 67.8 & 6.7 & 3.3 & 37.5 & 28.9 & 28.8 & 49.0 & 11.1 & 84.4 & 34.2 & 44.7\\
\pz\pz Qwen2.5-7B-Instruct~\cite{yang2024qwen25} & 75.5 & 10.0 & 6.7 & 52.5 & 35.5 & 36.0 & 53.0 & \textbf{58.1} & 91.3 & 43.2 & 61.4 \\
\pz\pz Qwen2.5-Math-7B-Instruct~\cite{yang2024qwen25mathtechnicalreportmathematical} & 83.6 & 16.7 & 10.0 & 62.5 & 41.6 & 42.9 &  51.3 & 28.0 & 85.3 & 36.2 & 50.2\\
\pz\pz rStar-Math-7B~\cite{guan2025rstar} & 78.4 & \textbf{26.7} & - & 47.5 & \textbf{47.1} & - & - & - & - & - & - \\
\pz\pz Eurus-2-7B-PRIME~\cite{cui2025process} & 80.4 & \textbf{26.7} & 13.3 & 60.0 & 43.7 & 44.8 & - & - & - & - & -\\
\grayrow
\multicolumn{12}{l}{\textit{Ours}} \\




\grayrow
\pz\pz \textbf{\method-Llama-8B} &70.4   &13.3 & 10.0 &50.3 & 36.0 & 36.0 &50.2 & 46.0 &88.2 & 32.3 &54.2 \\
\grayrow
\pz\pz \textbf{\method-Qwen-7B} & \textbf{84.6} & \textbf{26.7} & \textbf{23.3} & \textbf{75.0} & 46.1 & \textbf{51.1} & \textbf{66.8} & 57.0 & 92.0 & \textbf{43.8} & \textbf{64.9} \\

\bottomrule[1pt]
\end{tabular}}
\end{center}
\end{table*}

\begin{figure}[!t]
    \centering
    \captionsetup[subfigure]{justification=centering,singlelinecheck=false}
    \begin{subfigure}[b]{0.23\textwidth}
        \centering
        \includegraphics[width=\linewidth]{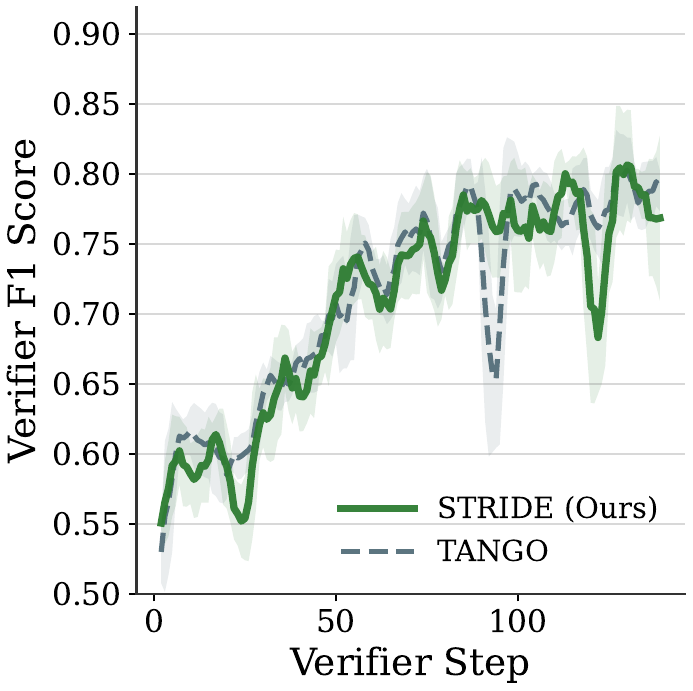}
        \caption{Verifier $F_1$}
    \end{subfigure}
    \hfill
    \begin{subfigure}[b]{0.28\textwidth}
        \centering
        \includegraphics[width=\linewidth]{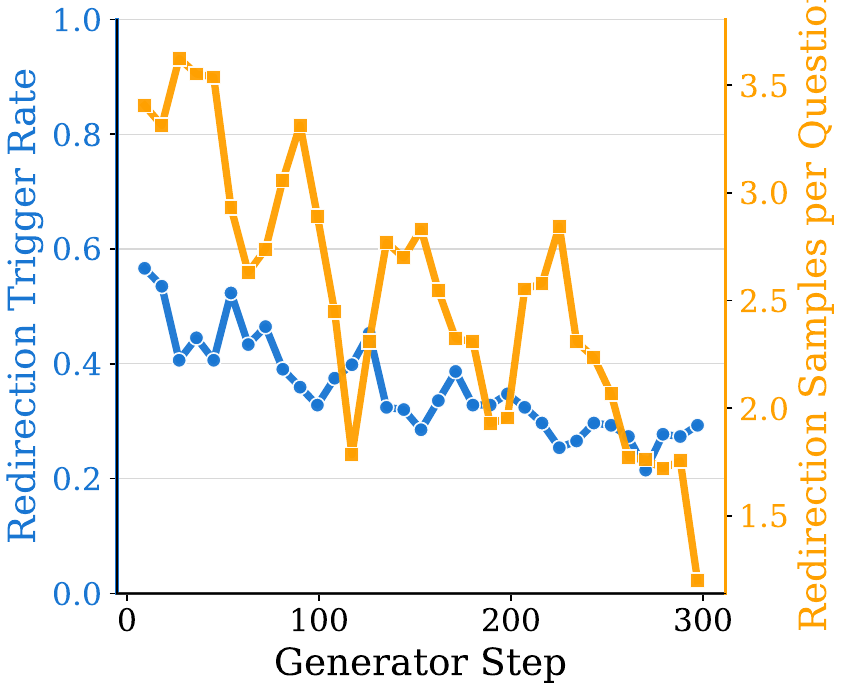}
        \caption{Guidance Efficiency}
    \end{subfigure}
    \hfill
    \begin{subfigure}[b]{0.23\textwidth}
        \centering
        \includegraphics[width=\linewidth]{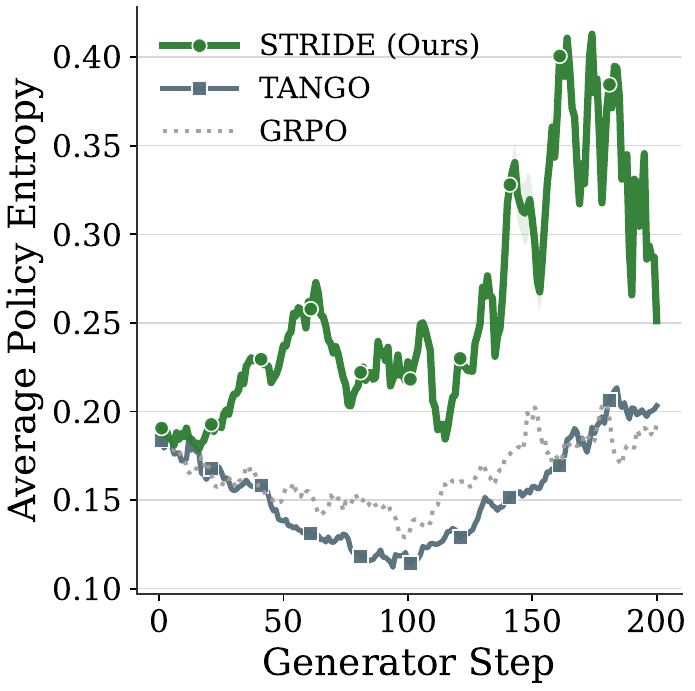}
        \caption{Policy Entropy}
    \end{subfigure}
    \hfill
    \begin{subfigure}[b]{0.23\textwidth}
        \centering
        \includegraphics[width=\linewidth]{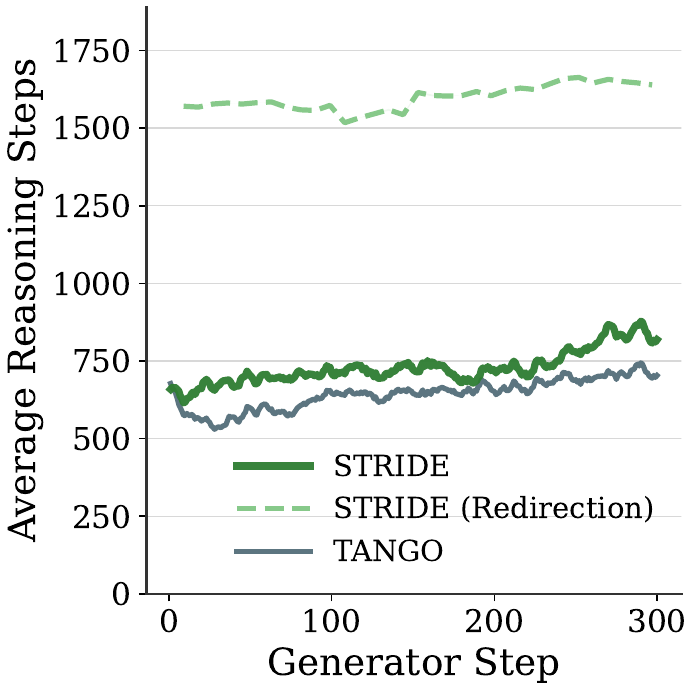}
        \caption{Reasoning Depth}
    \end{subfigure}
    \caption{\textbf{\method training dynamics.}
    (a)~Fair Comparison Validated: \method and TANGO share near-identical verifier $F_1$ trajectories, confirming the performance gap originates from \emph{how} feedback is utilized (language guidance vs.\ scalar reward).
    (b)~Continuous Breakthrough on Hard Problems: The declining redirection error rate shows the generator progressively conquers previously unsolvable instances, with the verifier pinpointing failures at ever-earlier steps as training matures.
    (c)~Sustained Exploration: \method maintains consistently higher policy entropy than TANGO, demonstrating that language guidance prevents premature convergence and sustains a richer exploration landscape throughout training.
    (d)~Emergence of Deep Reasoning: Redirected trajectories grow substantially longer during Phase~III, reflecting qualitatively richer reasoning chains that the generator could not produce through independent sampling alone.}
    \label{fig:main_results}
\end{figure}

\textbf{Evaluation Metrics.} We employ the zero-shot Pass@1 accuracy as our primary metric, using greedy decoding for all models. Furthermore, to specifically isolate the impact of our Phase III mechanism, we introduce the Correction Success Rate (CSR), which measures the probability of a generator successfully reaching the correct outcome after receiving a redirection trigger compared to its initial failed attempt. For the verifier, we measure its F1 score on verification accuracy to ensure its reliability. More implementation details are provided in the Appendix~\ref{app:implementation}.

\subsection{Main Results}
\textbf{Overall performance across domains.}
\Cref{tab:main} reports zero-shot Pass@1 results on both mathematics and out-of-domain reasoning benchmarks.
Across the 7B/8B scale models of various families, \method achieves the strongest overall performance, with consistent gains on nearly all tasks.
Crucially, these gains extend beyond math to logic, code, commonsense, and tabular reasoning, indicating that the generalization of \method is not confined to math domain but rather reflects consistent improvements in reasoning capabilities.

\textbf{Comparison with RLVR and co-training baselines.}
To isolate the effect of replacing scalar step rewards with stepwise redirection, we compare \method against vanilla outcome-based RLVR (GRPO) and the scalar-reward co-training baseline TANGO under identical settings.
As shown in \Cref{tab:rl_algorithms}, \method consistently outperforms both baselines on the two series across the two benchmark groups.
These results validate our claim: high-bandwidth linguistic guidance provides more actionable supervision than unidimensional scalar rewards, especially for multi-step reasoning where credit assignment is challenging.

\textbf{Training dynamics and the role of guidance.}
\Cref{fig:main_results} provides a closer look into the training process across four dimensions.
First, both \method and TANGO exhibit near-identical verifier $F_1$ growth curves (a), establishing that the two systems operate with comparable verifier quality throughout co-training.
This isolates the performance advantage of \method to the \emph{paradigm difference}: language guidance versus scalar step rewards, rather than a superior verifier.
Second, the simultaneous decline in redirection trigger rate and per-question sample yield (b) reveals maturing guidance efficiency: as training progresses, the generator fails on fewer problems, and the verifier localizes errors at increasingly earlier steps, reducing the cost and expanding the scope of each redirection cycle.
Third, this high-bandwidth supervision sustains consistently higher policy entropy in \method than in TANGO (c), indicating that stepwise language feedback preserves a broader exploration landscape and actively prevents the representational collapse observed in scalar-reward baselines.
Finally, the redirection phase fosters a qualitative shift in reasoning depth (d): trajectories generated under verifier guidance are substantially longer and more structurally complex than those produced by independent sampling, reflecting the emergence of deliberate, multi-step reasoning patterns that scalar supervision cannot elicit. These dynamics confirm the benefits of learnable language feedback.

\begin{table*}[!t]
\setlength{\tabcolsep}{4pt}
\renewcommand{\arraystretch}{1.1}
\caption{\small \textbf{Comparison of \methodbf with vanilla RLVR and co-training baselines}. By shifting process supervision from unidimensional rewards to stepwise redirection, \methodbf significantly outperforms GRPO and TANGO on both model series across mathematical reasoning and out-of-domain reasoning benchmarks. The baseline results of Qwen2.5 series are adopted from~\cite{zha2025rltango}. All models are trained for 200 generator steps with identical settings.}
\vspace{-0.2cm}
\label{tab:rl_algorithms}
\small
\begin{center}
{\small
\resizebox{\linewidth}{!}{
\begin{tabular}{lccccccccccc}
\toprule[1pt]
& \multicolumn{6}{c}{\textbf{Mathematical Reasoning}} & \multicolumn{5}{c}{\textbf{Out-of-Domain Reasoning}}  \\
\cmidrule(lr){2-7} \cmidrule(lr){8-12} \\[-1.0em]
\textbf{Model} & \textbf{MATH500} & \textbf{AIME2024} & \textbf{AIME2025} & \textbf{AMC2023}  & \textbf{OlympiadBench} & \textbf{Avg.} & \textbf{BGQA} & \textbf{CRUXEval} & \textbf{StrategyQA} & \textbf{TableBench} & \textbf{Avg.} \\ \midrule\midrule
Qwen2.5-7B-SFT & 66.6 & 3.3 & 3.3 & 27.5 & 28.1 & 25.8 & 46.6 & 44.3 & 85.9 & 34.4 & 52.8\\ 
\midrule
+ GRPO~\cite{shao2024deepseekmath_grpo} & 74.6 & 13.3 & 10.0 & 50.0 & 36.9 & 37.0 & 55.3 & 48.8 & 88.1 & 38.2 & 57.6 \\
+ TANGO~\cite{zha2025rltango}  & 81.4 & 20.0 & 20.0 & 65.0 & 43.9 & 46.1 & 60.5 & 51.4 & 90.0 & 42.3 & 61.1  \\
\grayrow
+ \methodbf~(Ours) & \textbf{84.6} & \textbf{26.7} & \textbf{23.3} & \textbf{75.0} & \textbf{46.1} & \textbf{51.1} & \textbf{66.8} & \textbf{57.0} & \textbf{92.0} & \textbf{43.8} & \textbf{64.9}  \\ 

\midrule
Llama3.1-8B-SFT &57.2 &6.6 &3.3 &42.5 &28.0 &27.5 & 46.5 & 43.5 &83.7 & 27.8 &50.4 \\ 
\midrule
+ GRPO~\cite{shao2024deepseekmath_grpo} &66.8 & 6.6 &6.6 & 47.5 & 34.2 & 32.3 & 48.2 & 45.5 & 84.6 & 30.8 & 52.3  \\
+ TANGO~\cite{zha2025rltango} &69.2 &6.6 &6.6 & 50.0 &35.0 & 33.5 & 49.0 & 45.3 &86.0 &28.9 &52.3  \\
\grayrow
+ \methodbf~(Ours) &\textbf{70.4}   &\textbf{13.3} & \textbf{10.0} &\textbf{50.3} & \textbf{36.0} & \textbf{36.0} &\textbf{50.2} & \textbf{46.0} &\textbf{88.2} & \textbf{32.3} &\textbf{54.2}  \\ 

\bottomrule[1pt]
\end{tabular}}}
\end{center}
\vspace{-1em}
\end{table*}
\subsection{Ablation Study}
\textbf{Core findings.}
\Cref{tab:strategy_ablation} decomposes Phase~III into two orthogonal choices: (i) \emph{anchor selection} (single-point $t^*$ vs. multi-point $[1,t^*]$) and (ii) \emph{supervision bandwidth} (no guidance vs. stepwise linguistic guidance).
Two conclusions stand out.
First, both factors are independently beneficial: multi-point anchors improve robustness to imperfect failure localization and capture earlier suboptimal decisions, while linguistic guidance provides actionable directions that scalar feedback cannot convey.
Second, their combination yields the largest gain, indicating that effective trajectory redirection requires \emph{both} reliable restart states and high-bandwidth corrective signals.
Finally, we find that injecting scalar step rewards into \method can hurt performance compared to our outcome-only design, suggesting that misaligned step rewards may introduce noise and destabilize learning.
Critically, \method achieves a CSR of 6.8\% on zero-pass-rate problems, where all scalar-reward baselines yield identically zero gradient signal, directly quantifying Phase~III's role as a breakthrough mechanism that converts otherwise unsolvable problems into productive training signals.

\begin{wraptable}{r}{0.60\linewidth}
\vspace{-1em}
\centering
\small
\caption{\textbf{Ablation of Redirection Strategies.} This table evaluates the contribution of anchor selection and linguistic guidance on MATH-500 with performance and CSR. *Single-Point denotes no linguistic guidance with only a single-point anchor at $t^*$. *\method indicates including step-level reward for training. Results confirm that combining multi-point anchors with high-bandwidth guidance yields the most significant performance breakthrough.}
\label{tab:strategy_ablation}
\begin{tabularx}{\linewidth}{@{}Xcccc@{}}
\toprule
\textbf{Configuration} & \textbf{Anchor} & \textbf{Guidance} & \textbf{MATH} & \textbf{CSR} \\ \midrule
(i) Standard RLVR & None & None & 74.6 & - \\
(ii) *Single-Point  & $t^*$ only & None & 74.8 & 1.2 \\
(iii) Single-Point & $t^*$ only & Ling. & 78.5 & 3.4 \\
(iv) Multi-Point & $[1, t^*]$ & None & 79.1 & 3.8 \\
(v) *\method & $[1, t^*]$ & Ling. & 82.6 &  5.2 \\
(vi) \methodbf & $[1, t^*]$ & \textbf{Ling.} & \textbf{84.6} & \textbf{6.8} \\
\bottomrule
\end{tabularx}
\vspace{-1em}
\end{wraptable}

\begin{figure}[b]
    \centering
    \captionsetup[subfigure]{justification=centering,singlelinecheck=false}
    \begin{subfigure}[b]{0.22\linewidth}
        \centering
        \includegraphics[width=\textwidth]{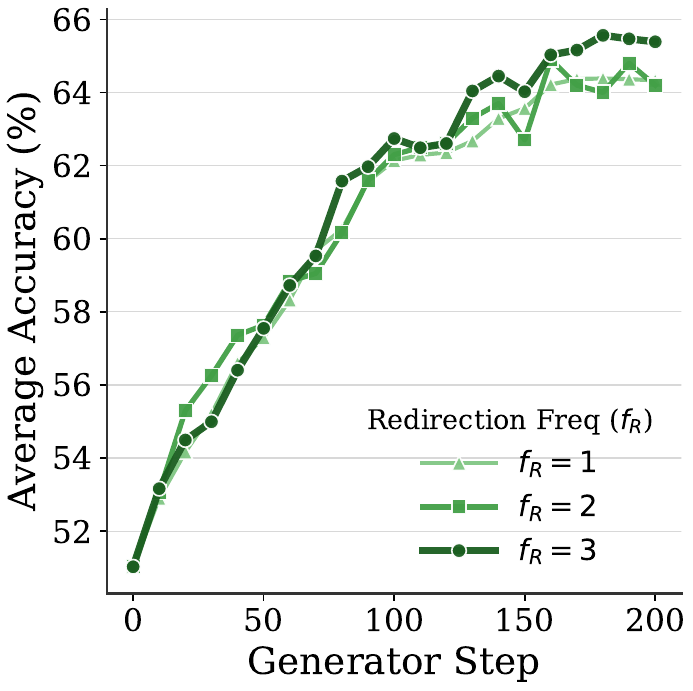}
        \caption{Redirection\\Frequency ($f_R$)}
        \label{fig:freq_ablation}
    \end{subfigure}
    \hfill
    \begin{subfigure}[b]{0.22\linewidth}
        \centering
        \includegraphics[width=\textwidth]{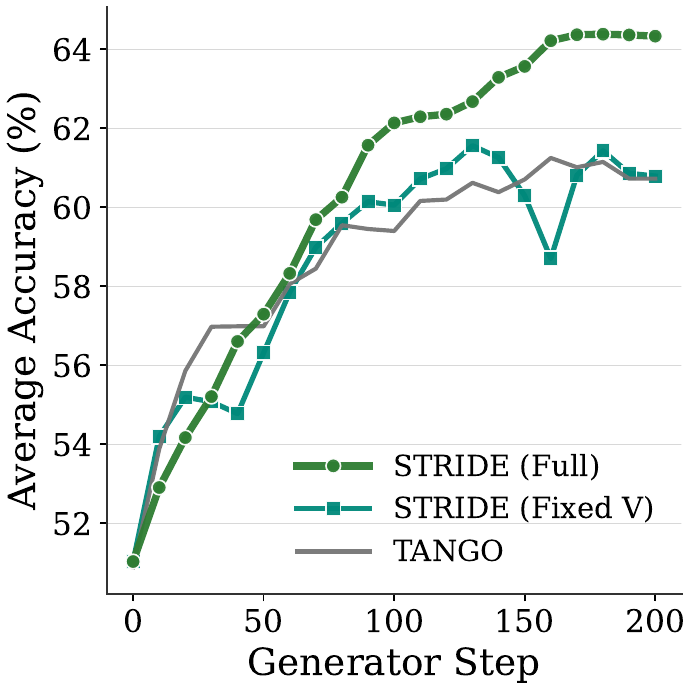}
        \caption{Impact of\\Co-training}
        \label{fig:reward_ablation}
    \end{subfigure}
    \hfill
    \begin{subfigure}[b]{0.25\linewidth}
        \centering
        \includegraphics[width=\textwidth]{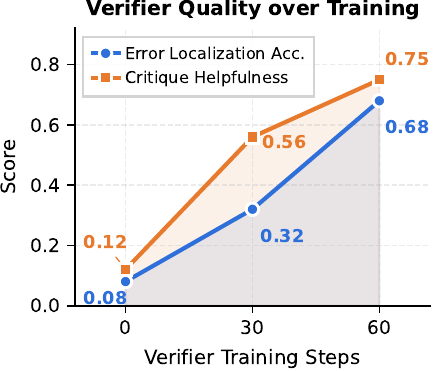}
        \caption{Verifier Quality\\over Training}
        \label{fig:verifier_quality}
    \end{subfigure}
    \hfill
    \begin{subfigure}[b]{0.25\linewidth}
        \centering
        \includegraphics[width=\textwidth]{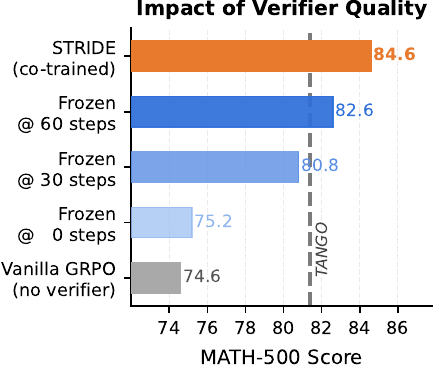}
        \caption{Impact of\\Verifier Quality}
        \label{fig:frozen_verifier}
    \end{subfigure}
    \caption{\textbf{Sensitivity Analysis and Verifier Quality Assessment.}
    (a) Higher redirection frequency ($f_R=1,2,3$) yields marginal but consistent Pass@1 gains.
    (b) Co-training the verifier and generator is important for maintaining high-quality guidance signals.
    (c) Verifier step-level localization accuracy and critique helpfulness both improve steadily over training, confirming that outcome-level supervision yields reliable process-level capabilities.
    (d) Higher verifier quality at the time of freezing yields consistently better MATH-500 performance, while co-training (STRIDE) achieves the strongest result, validating the importance of joint optimization.}
    \label{fig:ablation_sensitivity}
    \vspace{-1em}
\end{figure}

\textbf{Sensitivity and Verifier Quality Assessment.}
\Cref{fig:freq_ablation} shows that higher redirection frequency $f_R$ yields marginal but consistent Pass@1 gains, suggesting Phase~III acts as a ``rare-but-high-value'' correction operator whose returns saturate once easy-to-correct failures are exhausted.
\Cref{fig:reward_ablation} further confirms that co-training the verifier with the generator is crucial: the fixed-verifier variant underperforms co-trained \method, as a frozen verifier cannot adapt its localization to the generator's evolving error distribution.

To directly characterize verifier reliability, \Cref{fig:verifier_quality} tracks step-level quality over training using GPT-5 as an automatic judge, measuring error localization accuracy and critique helpfulness independently.
Both metrics improve substantially (localization: $0.08 \to 0.68$; helpfulness: $0.12 \to 0.75$), confirming that outcome-level RL supervision is sufficient to develop reliable process-level capabilities, analogous to how chain-of-thought reasoning emerges from outcome reward alone.

\Cref{fig:frozen_verifier} further isolates the impact of verifier quality by comparing \method against variants that freeze the verifier at different training stages.
Performance degrades monotonically as verifier quality drops, yet even the weakest frozen verifier (0 steps) still outperforms vanilla GRPO ($75.2$ vs.\ $74.6$).
This confirms that \method's Multi-Point Redirection Strategy provides built-in robustness to imperfect verification: even when localization is unreliable, the outcome-only reward ensures that only successful redirections contribute gradient updates, guaranteeing non-negative improvement by design. This robustness also demonstrates that paradigm shifts toward language feedback can yield benefits from imperfect verifiers, while noisy of scalar rewards can degrade performance, alleviating concerns about verifier reliability and effectively leverage verifier feedback for policy improvement, even when the verifier is still learning
in early training stages.

\section{Conclusion}
We propose \method, an interleaved three-phase training framework that shifts process supervision in RLVR from sparse scalar rewards to high-bandwidth stepwise language feedback.
By co-training a generative verifier with outcome-only rewards and using its language critiques to trigger guided trajectory redirection, \method alleviates the information bottleneck of scalar supervision and turns failed trajectories into efficient learning signals.
Extensive experiments across mathematical and out-of-domain reasoning benchmarks demonstrate consistent improvements over outcome-only RLVR and scalar-reward co-training baselines, while ablations confirm the importance of multi-point anchors and linguistic guidance. A promising direction for future work is to extend the redirection mechanism to more general agentic settings, which involve multi-step tool calls.


\clearpage

\section*{Acknowledgments}
This research is supported by the RIE2025 Industry Alignment Fund – Industry Collaboration Projects (IAF-ICP) (Award I2301E0026), administered by A*STAR, as well as supported by Alibaba Group and NTU Singapore through Alibaba-NTU Global e-Sustainability CorpLab (ANGEL).

\bibliography{example_paper}
\bibliographystyle{unsrt}

\newpage
\appendix
\part*{Appendix}
\vspace*{20pt}
\section*{Table of Contents}
\hypersetup{
  linkcolor=darkblue   
}
\startcontents[sections]
\printcontents[sections]{l}{1}{\setcounter{tocdepth}{2}}
\hypersetup{
  linkcolor=red   
}

\newpage

\section{STRIDE Interleaved Training Algorithm}
\label{app:algorithm}

\begin{algorithm}[H]
\caption{\method Interleaved Training}
\label{alg:interleaved}
\begin{algorithmic}[1]
\STATE \textbf{Initialize:} Generator $G_\theta$, Verifier $V_\phi$, Iterations $N$
\STATE \textbf{Set Frequencies:} $f_G = 9, f_V = 3, f_R = 1$
\FOR{each training cycle $C = 1, \dots, N$}
    \STATE \textcolor{c4}{\textbf{// Phase I: Base Policy Optimization}}
    \REPEAT
        \STATE Sample $y \sim G_\theta(x)$; Update $\theta$ via GRPO with outcome $c_O^*$.
    \UNTIL{run for $f_G$ steps, with Phase II injected every 3 $G$-steps}

    \STATE \textcolor{c3}{\textbf{// Phase II: Generative Verifier Optimization}}
    \STATE Sample $y \sim G_\theta(x), v \sim V_\phi(x, y)$; Update $\phi$ to produce stepwise verification $v$ via GRPO with $r = \mathbb{I}(\hat{c_O} = c_O^*)$

    \STATE \textcolor{c2}{\textbf{// Phase III: Guided Trajectory Redirection}}
    \STATE \textbf{Execute once per cycle ($f_R=1$):}
    \STATE 1. \textbf{Selection:} Filter queries $x$ where $G_\theta$ failed all attempts ($c_O^* = 0$) and $V_\phi$ correctly identified the failure ($c_O= 0$).
    \STATE 2. \textbf{Error Localization:} Identify first point of failure $t^* = \min \{ t \mid \tau(v_t) = 0 \}$.
    \STATE 3. \textbf{Parallel Reconstruction:} Build $\{S_{redirect}^{(t)}\}_{t=1}^{t^*}$ anchors using prefix $(x, z_{<t}, v_{<t})$.
    \STATE 4. \textbf{Redirection:} Sample $y_{red} \sim G_\theta(S_{redirect})$ and update $\theta$ via GRPO with outcome $c_O^*$.
\ENDFOR
\end{algorithmic}
\end{algorithm}

\section{Formal Preliminaries}
\label{app:prelim_formal}

\subsection{Generator-Verifier Framework}

\textbf{Generator as Reasoner.}
Given an input query $x \in \mathcal{X}$, the generator $G_\theta$ aims to generate a reasoning path $y$ consisting of a sequence of intermediate thought steps, denoted as $y = (z_1, z_2, \dots, z_T) \in \mathcal{Y}$. The probability of generating a specific path is factorized as:
\begin{equation}
    G_\theta(y | x) = \prod_{t=1}^T G_\theta(z_t | x, z_{<t}),
\end{equation}
where $\mathcal{Y}$ represents the high-dimensional discrete space of natural language reasoning.

\textbf{Verifier as Evaluator.}
The verifier $V_\phi$ assesses the correctness $c_t \in \{0, 1\}$ of each step $z_t$ and the overall correctness $c_O \in \{0, 1\}$ of path $y$ by producing a generative verification sequence $v = (v_1, v_2, \dots, v_T) = V_\phi(x, y),\ v \in \mathcal{V}^*$, where $\mathcal{V}^*$ denotes the discrete space of natural language. Step-level scores $(\hat{c}_1, \dots, \hat{c}_T)$ and an overall judgment $\hat{c}_O$ are parsed from this sequence. The verifier is trained via RLVR with outcome-based supervision: the reward is 1 if $\hat{c}_O = c_O^*$, and 0 otherwise.

\textbf{Optimization with Step-level Reward.}
In the original GV framework, the generator is updated with a mixed advantage:
\begin{equation}
    \hat{A}_t = (1 - \alpha)\hat{A}_{t,\text{out}} + \alpha\hat{A}_{t,\text{step}},
\end{equation}
where $\alpha \in (0,1)$ decays exponentially to shift focus from step-level to outcome-based supervision. The resulting optimization objective is:
\begin{equation}
    \mathcal{J}(\theta) = \mathbb{E}_{x, y \sim G_\theta} \left[ \sum_{t=1}^T \nabla_\theta \log G_\theta(z_t | x, z_{<t}) \cdot \hat{A}_t \right].
\end{equation}
While this mitigates supervision sparsity, the step-level reward from the verifier is not easily aligned with the ground-truth outcome signal, creating a critical instability in training.

\subsection{Information Bottleneck: Rate-Distortion Analysis}
\label{app:ib_formal}

We formalize the information bottleneck of scalar rewards via Rate-Distortion theory. Let the mapping $f: \mathcal{Y} \to \mathbb{R}$ compress reasoning paths to scalar rewards, and let $Z$ denote the latent oracle guidance representing ground-truth logical steps. According to Rate-Distortion theory~\cite{cover1999elements}, the mutual information $I(R; Z)$, quantifying the effective guidance provided by the reward, is bounded by the entropy of the reward signal:
\begin{equation}
    I(R; Z) \leq H(R).
\end{equation}
For a binary scalar reward $r \in \{0,1\}$, $H(R) \leq 1$ bit. In contrast, the complexity of $\mathcal{Y}$ requires a substantially higher bitrate to uniquely identify and rectify diverse logical fallacies.

Furthermore, since $\text{dim}(\mathcal{Y}) \gg \text{dim}(\mathbb{R})$, $f$ is \textbf{heavily many-to-one (non-injective)}: for a given $r$, the pre-image $f^{-1}(r) = \{y \in \mathcal{Y} \mid f(y) = r\}$ contains a vast number of semantically distinct paths. Paths with fundamentally different logical errors may receive identical scalar values, and the resulting gradient signal is insufficient to distinguish \textit{where} or \textit{why} a mistake occurred.

\subsection{Illustrative Example: Representational Collapse}
\label{app:rep_collapse}

\begin{motivationbox}{Representational Collapse in Scalar Rewards}
\textbf{Question:} Solve for $x$ in $2x + 5 = 13$. \\
\textbf{Previous Step:} $2x + 5 = 13$ (Initial state).

\vspace{0.8em}
\textit{The mapping from reasoning paths to rewards is \textbf{heavily non-injective}. As shown below, disparate error modes are collapsed into identical scalar values, providing no discriminative guidance.}

\vspace{0.5em}
\small
\begin{tabularx}{\linewidth}{
    >{\hsize=1.0\hsize\raggedright\arraybackslash}X
    >{\hsize=0.6\hsize\centering\arraybackslash}X
    >{\hsize=1.4\hsize\raggedright\arraybackslash}X
}
    \toprule
    \textbf{Current Erroneous Step} & \textbf{Scalar Reward} & \textbf{Language Feedback} \\
    \midrule
    \textit{Case A \newline (Arithmetic Error):} \newline
    $2x = 13 + 5$ &
    $r = 0.2$ \newline \color{red}\faTimes \space \footnotesize (Low) &
    \textbf{Incorrect:} \newline Incorrect additive inverse applied during transposition of $+5$. \\
    \addlinespace[0.8em]
    \textit{Case B \newline (Logical Leap):} \newline
    $2x + 5 = 13 \implies x = 3$ &
    $r = 0.2$ \newline \color{red}\faTimes \space \footnotesize (Low) &
    \textbf{Incorrect:} \newline Intermediate algebraic transformations are missing between the premise and result. \\
    \bottomrule
\end{tabularx}
\vspace{0.5em}

\textbf{Conclusion:} Scalar rewards suffer from \textit{representational collapse}, providing a magnitude without direction. Language feedback breaks this bottleneck by restoring the semantic dimensions of feedback.
\end{motivationbox}

\section{Prompt Templates \& Redirection Instructions}
\label{app:prompts}
To ensure a high-bandwidth information flow and maintain structural consistency, \method employs standardized prompt templates for reasoning, verification, and trajectory redirection. This section details the specific instructions provided to the Generator ($G_{\theta}$) and the Verifier ($V_{\phi}$) across the three training phases.
\subsection{Phase I Generator Reasoning Template}
In Phase I, the generator is tasked with basic stepwise reasoning. The template enforces a strict \texttt{<think>} and \texttt{<step>} structure to facilitate error localization in subsequent phases.
\begin{tcolorbox}[title=Generator Reasoning Template ($G_{\theta}$), colback=gray!5, colframe=blue!75!black,
    fonttitle=\bfseries,
    arc=4pt,
    outer arc=4pt]
You are a helpful Assistant that solves mathematical problems step-by-step. \\
You MUST follow this exact format:
\begin{enumerate}
    \item Start with a \texttt{<think>} section containing your step-by-step reasoning.
    \item Inside \texttt{<think>}, each distinct logical step MUST be enclosed in its own \texttt{<step> </step>} tags.
    \item After \texttt{<think>}, provide the final answer within \texttt{<answer> </answer>} tags, using the \textbackslash{}boxed\{\} format.
\end{enumerate}
... [Detailed Example Omitted] ... \\
User: {{prompt}} \\
Assistant:
\end{tcolorbox}

\subsection{Phase II Generative Verification Template}
The verifier $V_{\phi}$ is optimized in Phase II to act as a Contextual Navigator. It decomposes the terminal outcome signal into high-bandwidth linguistic feedback for each reasoning step.

\begin{tcolorbox}[
    title=Generative Verification Template ($V_{\phi}$),
    colback=green!2,
    colframe=green!40!black,
    fonttitle=\bfseries,
    arc=4pt,
    outer arc=4pt
]
\small
\textbf{System Instruction:} \\
You are a verification assistant specialized in mathematical reasoning. Your task is to carefully evaluate the provided solution step by step, checking for mathematical correctness and logical coherence... You MUST verify EACH \texttt{<step>} block found in the Assistant's solution.

\vspace{0.5em}
\textbf{Problem:} \{problem\} \\
\textbf{Assistant's Solution:} \{solution\} \\
\textbf{Step Count:} The Assistant's solution contains \{generator\_step\_count\} steps.

\vspace{0.5em}
\textbf{Task Guidance:} \\
For each of the \{generator\_step\_count\} steps, you MUST provide ONE corresponding verification analysis within a \texttt{<step>} tag inside the \texttt{<step\_verification>} section.

\vspace{0.5em}
\textbf{Required Format:} \\
\texttt{<step\_verification>} \\
\texttt{<step>}Step 1 Analysis: [Your reasoning]. \textbackslash{}boxed\{CORRECT/INCORRECT\}\texttt{</step>} \\
... \\
\texttt{</step\_verification>} \\
\texttt{<final\_verification>}\textbackslash{}boxed\{CORRECT/INCORRECT\}\texttt{</final\_verification>}

\vspace{0.5em}
\textbf{Constraint Checklist:}
\begin{itemize}
    \item Analyze EVERY step individually.
    \item Provide original reasoning, do not copy solution text.
    \item Each step analysis must end with exactly one \textbackslash{}boxed{} judgment.
    \item Output ONLY the specified tags.
\end{itemize}
\end{tcolorbox}

\subsection{Phase III Redirection Instructions}
In Phase III, \method distinguishes between Rectification and Exploration to maximize the utility of failure cases.
\begin{itemize}
    \item Rectification Prompt: Triggered at the First Point of Failure (FPF, $t=t^*$), providing explicit feedback to correct the detected logical fallacy.
    \item Exploration Prompt: Triggered at pre-failure anchors ($t < t^*$), encouraging the discovery of alternative robust paths from verified semantic anchors.
\end{itemize}

\begin{tcolorbox}[
    title=Redirection: Rectification (at FPF),
    colback=orange!2,
    colframe=orange!50!black,
    fonttitle=\bfseries
]
\small
\textbf{System Instruction:} \\
You are a mathematical Assistant. You will be shown a problem, a previous attempt, and step-by-step verification feedback. Your task is to correct the last step of previous attempt and continue solving the problem.

\vspace{0.5em}
\textbf{Problem:} \{problem\} \\
\textbf{[Previous Attempt \& Feedback]:} \\
\texttt{\{combined\_body\}} 

\vspace{0.5em}
\textbf{Task Guidance:} \\
The verification detected the error in the last step. Now you'll correct the error and continue answering.

\vspace{0.5em}
\textbf{Your Solution:}
\end{tcolorbox}

\begin{tcolorbox}[
    title=Redirection: Exploration (before FPF),
    colback=purple!2,
    colframe=purple!50!black,
    fonttitle=\bfseries
]
\small
\textbf{System Instruction:} \\
You are a mathematical Assistant. You will be shown a problem, a previous attempt, and step-by-step verification feedback. Your task is to continue solving the problem from the last step of previous attempt.

\vspace{0.5em}
\textbf{Problem:} \{problem\} \\
\textbf{[Previous Attempt \& Feedback]:} \\
\texttt{\{combined\_body\}} 

\vspace{0.5em}
\textbf{Task Guidance:} \\
The verification guided the previous steps. Now you'll continue answering.

\vspace{0.5em}
\textbf{Your Solution:}
\end{tcolorbox}

\section{Implementation Details} 
\label{app:implementation}
Our training framework is implemented using the \textbf{veRL}~\cite{sheng2024hybridflow_verl_codebase} distributed RL library and follows the interleaved schedule described in \Cref{sec:method}. We set the cadence ratio to $f_G:f_V:f_R = 9:3:1$, meaning the verifier updates once every three generator steps to compensate for the relative complexity of policy optimization, while the redirection phase (Phase III) is activated once per full cycle.

\textbf{Data Preparation and SFT.}
Consistent with prior art~\cite{zha2025rltango}, we perform initial Supervised Fine-Tuning (SFT) on the generator using 113K competition-level math prompts from the \textit{Eurus-2-SFT-Data}~\cite{cui2025process}. To ensure the data quality, reasoning trajectories are generated by prompting \texttt{Llama-3.1-70B-Instruct} with a decoding temperature of 0.1 and top-$p$ of 0.5, enforcing the step-by-step reasoning format within \texttt{<step>} tags. We use a full-parameter SFT with a learning rate of $5 \times 10^{-6}$ and a cosine annealing schedule. For Qwen2.5 we use the model after 800 SFT steps, and for Llama-3.1 we use the model after 1,000 SFT steps as the base generator $G_\theta$ for subsequent RL training. Notably, the verifier is initialized directly from the base model without prior SFT to demonstrate the framework's capability to bootstrap from a weaker starting point through mutual reinforcement.

\textbf{Reinforcement Learning Configurations.}
During the RL stage, we employ 455K question-answer pairs from \textit{Eurus-2-RL-Data}~\cite{cui2025process}. All training is conducted using the \textbf{GRPO} algorithm with a group size of $M=5$ rollouts per prompt. To prevent early training instability, we implement a \textbf{verifier warmup} of 40 steps, allowing the verifier to learn output formatting and basic correctness before providing redirection guidance to the generator. We use a constant learning rate of $1 \times 10^{-6}$, a total batch size of 256, and a KL-divergence penalty coefficient $\beta = 0.001$. For the ablation fixed-verifier setting, we freeze $V_\phi$ after the warmup phase to isolate the effect of joint training.

\textbf{Stability and Robustness.}
A key advantage of \method is its inherent stability without exhaustive hyperparameter tuning. Unlike discriminative PRMs that require complex reward mixing, \method maintains \textbf{outcome-only reward grounding}. The advantage in Phase III is tied strictly to ground-truth outcome correctness, which shields the policy from potential verifier hallucinations or noisy step-level rewards. This design choice simplifies training dynamics and enhances robustness, as only successful redirections contribute to policy updates in GRPO.

\section{Additional Experimental Results}
\label{app:additional_exp}

\subsection{Compute Fairness Analysis}

A natural concern is whether \method's gains over GRPO stem from an increased computational budget rather than from the paradigm shift in process supervision.
To address this, we compare \method against compute-enhanced GRPO baselines that match \method's total GPU hours (approximately 40 hours on 8$\times$H20 GPUs) by either increasing training steps or enlarging the rollout batch size.
As shown in \Cref{tab:compute_fairness}, both compute-enhanced GRPO variants yield negligible gains over vanilla GRPO (74.4 and 75.2 vs.\ 74.6), while \method achieves 84.6 under the same budget.
This confirms that the improvement originates from the quality of supervision signals, not from additional compute.

\begin{table}[h]
\centering
\small
\caption{\textbf{Compute Fairness Analysis.} All methods are evaluated under matched GPU hours ($\approx$40 hours on 8$\times$H20 GPUs). Compute-enhanced GRPO baselines with more training steps or larger rollout batches yield negligible gains over vanilla GRPO, confirming that \method's improvement stems from the paradigm shift in process supervision rather than increased computational budget.}
\label{tab:compute_fairness}
\setlength{\tabcolsep}{6pt}
\begin{tabular}{lccc}
\toprule
\textbf{Method} & \textbf{Training Time (hr)} & \textbf{RL Steps} & \textbf{MATH-500} \\
\midrule
Vanilla GRPO            & 32 & 200 & 74.6 $\pm$ 0.2 \\
GRPO (more steps)       & 40 & 250 & 74.4 $\pm$ 0.2 \\
GRPO (bigger batch)     & 40 & 200 & 75.2 $\pm$ 0.5 \\
TANGO                   & 40 & 200 & 81.4 $\pm$ 0.3 \\
\midrule
\grayrow \methodbf      & 40 & 200 & \textbf{84.6} $\pm$ 0.2 \\
\bottomrule
\end{tabular}
\end{table}

\subsection{Orthogonality with Inference-Time Scaling}

\method is a training-time method and is by design orthogonal to inference-time scaling techniques.
To empirically verify this, we combine \method with Best-of-N (BoN) sampling and a Process Reward Model (PRM) and compare against GRPO under the same inference budget.
\Cref{tab:inference_scaling} shows two key findings.
First, \method's pass@1 result (84.6) already surpasses GRPO with $16\times$ more inference trajectories (79.6), demonstrating that training-time language guidance provides a fundamentally stronger policy than scalar-reward RL regardless of inference effort.
Second, \method further compounds with inference-time scaling: combining \method with BoN(8)+PRM achieves 88.2, substantially outperforming the corresponding GRPO+BoN(8)+PRM baseline (78.2).
These results confirm that training-time language guidance and inference-time search address orthogonal bottlenecks and are mutually beneficial.

\begin{table}[h]
\centering
\small
\caption{\textbf{Orthogonality with Inference-Time Scaling.} \method pass@1 surpasses GRPO with $16\times$ more inference trajectories, and further compounds with Best-of-N sampling and a PRM to achieve 88.2 on MATH-500, demonstrating that training-time language guidance and inference-time scaling are orthogonal and mutually beneficial.}
\label{tab:inference_scaling}
\setlength{\tabcolsep}{6pt}
\begin{tabular}{lccc}
\toprule
\textbf{Method} & \textbf{Training Time (hr)} & \textbf{Inference Traj.} & \textbf{MATH-500} \\
\midrule
GRPO (pass@1)            & 32 & 1  & 74.6 $\pm$ 0.2 \\
GRPO + BoN(8) + PRM      & 32 & 8  & 78.2 $\pm$ 0.3 \\
GRPO + BoN(16) + PRM     & 32 & 16 & 79.6 $\pm$ 0.4 \\
\midrule
\grayrow \methodbf\ (pass@1)         & 40 & 1  & \textbf{84.6} $\pm$ 0.2 \\
\grayrow \methodbf\ + BoN(8) + PRM   & 40 & 8  & \textbf{88.2} $\pm$ 0.3 \\
\bottomrule
\end{tabular}
\end{table}

\subsection{Results with Variance: Comparison with Baselines}
\label{app:variance_analysis}

\Cref{tab:rl_algorithms_std} reproduces the main comparison table with standard deviations reported on the two Avg.\ columns, computed across three independent runs with different random seeds.

\begin{table*}[h]
\setlength{\tabcolsep}{4pt}
\renewcommand{\arraystretch}{1.1}
\caption{\small \textbf{Comparison of \methodbf with vanilla RLVR and co-training baselines (with variance).} Standard deviations on Avg.\ columns are computed across three independent runs. Individual benchmark scores are reported as single-run results.}
\label{tab:rl_algorithms_std}
\small
\begin{center}
{\small
\resizebox{\linewidth}{!}{
\begin{tabular}{lccccccccccc}
\toprule[1pt]
& \multicolumn{6}{c}{\textbf{Mathematical Reasoning}} & \multicolumn{5}{c}{\textbf{Out-of-Domain Reasoning}}  \\
\cmidrule(lr){2-7} \cmidrule(lr){8-12} \\[-1.0em]
\textbf{Model} & \textbf{MATH500} & \textbf{AIME2024} & \textbf{AIME2025} & \textbf{AMC2023} & \textbf{OlympiadBench} & \textbf{Avg.} & \textbf{BGQA} & \textbf{CRUXEval} & \textbf{StrategyQA} & \textbf{TableBench} & \textbf{Avg.} \\ \midrule\midrule
Qwen2.5-7B-SFT & 66.6 & 3.3 & 3.3 & 27.5 & 28.1 & 25.8 $\pm$ 2.1 & 46.6 & 44.3 & 85.9 & 34.4 & 52.8 $\pm$ 2.4\\
\midrule
+ GRPO~\cite{shao2024deepseekmath_grpo} & 74.6 & 13.3 & 10.0 & 50.0 & 36.9 & 37.0 $\pm$ 1.8 & 55.3 & 48.8 & 88.1 & 38.2 & 57.6 $\pm$ 2.3\\
+ TANGO~\cite{zha2025rltango}           & 81.4 & 20.0 & 20.0 & 65.0 & 43.9 & 46.1 $\pm$ 2.0 & 60.5 & 51.4 & 90.0 & 42.3 & 61.1 $\pm$ 2.1\\
\grayrow
+ \methodbf~(Ours) & \textbf{84.6} & \textbf{26.7} & \textbf{23.3} & \textbf{75.0} & \textbf{46.1} & \textbf{51.1} $\pm$ 2.2 & \textbf{66.8} & \textbf{57.0} & \textbf{92.0} & \textbf{43.8} & \textbf{64.9} $\pm$ 2.3\\
\midrule
Llama3.1-8B-SFT & 57.2 & 6.6 & 3.3 & 42.5 & 28.0 & 27.5 $\pm$ 1.2 & 46.5 & 43.5 & 83.7 & 27.8 & 50.4 $\pm$ 1.4\\
\midrule
+ GRPO~\cite{shao2024deepseekmath_grpo} & 66.8 & 6.6 & 6.6 & 47.5 & 34.2 & 32.3 $\pm$ 1.7 & 48.2 & 45.5 & 84.6 & 30.8 & 52.3 $\pm$ 2.1\\
+ TANGO~\cite{zha2025rltango}           & 69.2 & 6.6 & 6.6 & 50.0 & 35.0 & 33.5 $\pm$ 2.0 & 49.0 & 45.3 & 86.0 & 28.9 & 52.3 $\pm$ 1.8\\
\grayrow
+ \methodbf~(Ours) & \textbf{70.4} & \textbf{13.3} & \textbf{10.0} & \textbf{50.3} & \textbf{36.0} & \textbf{36.0} $\pm$ 1.8 & \textbf{50.2} & \textbf{46.0} & \textbf{88.2} & \textbf{32.3} & \textbf{54.2} $\pm$ 2.0\\
\bottomrule[1pt]
\end{tabular}}}
\end{center}
\end{table*}

\section{Limitations}
\label{app:limitations}
\method introduces additional training complexity relative to vanilla RLVR, as it requires maintaining two separate models and executing an interleaved three-phase schedule; in our current implementation this amounts to approximately 40 hours on 8$\times$H20 GPUs, compared to 32 hours for standard GRPO.
The computational overhead could be reduced through selective redirection targeting only the most informative failure cases, parameter sharing between the generator and verifier, or parallelizing the three training phases.
Additionally, while \method is evaluated on mathematical and general reasoning benchmarks, its extension to open-ended tasks with non-verifiable outputs (e.g., creative writing or complex dialogue) would require rubric-based outcome signals such as LLM-as-a-judge, which introduces additional variance into the reward signal.
Finally, the verifier's step-level localization, while improving over training, is not perfect; future work on dedicated process-level regularization could further strengthen its reliability.

\section{Case Studies}
This section provides qualitative evidence demonstrating how \method overcomes the inherent limitations of scalar rewards through high-bandwidth linguistic guidance.

\subsection{Case Study I: Breaking Representational Collapse}
As discussed in \Cref{sec:ib_bottleneck}, a fundamental weakness of conventional PRMs is their unidimensional nature, which collapses semantically distinct errors into identical scalar values. \Cref{tab:case_collapse} showcases two disparate error modes in an algebraic problem that are indistinguishable to a scalar verifier but are clearly resolved by \method.

\begin{table}[h]
    \centering
    \small
    \caption{Comparison of feedback bandwidth between Scalar Rewards and \methodbf.}
    \label{tab:case_collapse}
    \begin{tabular}{p{0.28\linewidth}p{0.18\linewidth}p{0.45\linewidth}}\toprule
        \textbf{Error Mode} & \textbf{Scalar Reward} & \textbf{\methodbf Guidance (Informative Critique)} \\
        \midrule
        \textbf{Case A: Arithmetic Fallacy} \newline $2x+5=13 \Rightarrow 2x=18$ & $r=0.2$ (Low) & \textcolor{red}{Incorrect additive inverse applied.} The step incorrectly added 5 to the RHS instead of subtracting it. \\
        \midrule
        \textbf{Case B: Logical Leap} \newline $2x+5=13 \Rightarrow x=3$ & $r=0.2$ (Low) & \textcolor{red}{Missing intermediate steps.} While the result $x=4$ is intended, the jump to $x=3$ is both logically unsubstantiated and numerically wrong. \\
\bottomrule
\end{tabular}
\end{table}

In both cases, a scalar reward model provides only a magnitude of failure (e.g., 0.2), leaving the generator to explore blindly. In contrast, \methodbf restores the semantic dimension, providing a clear gradient direction in natural language for the generator to redirect its trajectory.

\subsection{Case Study II: Multi-Point Redirection Produces Diverse Strategies}
A key property of Multi-Point Redirection is that redirecting from \emph{different} prefix anchors elicits qualitatively different solution strategies, even for the same problem.
We trace two concurrent redirection paths on the same failed trajectory to illustrate this diversity.

\begin{tcolorbox}[title=Initial Attempt (Failed), fonttitle=\bfseries, colback=gray!3!white, colframe=gray!55!black, arc=2mm, boxrule=0.6pt]
\small
\textbf{Problem:} Solve for $x$: $2^{2x} - 5(2^x) + 4 = 0$.
\begin{itemize}
  \item \texttt{<step 1>} Let $u = 2^x$. Rewrite as $u^2 - 5u + 4 = 0$. \hfill \textit{Verified: CORRECT}
  \item \texttt{<step 2>} Factor: $(u-4)(u+1)=0$, so $u=4$ or $u=-1$. \hfill \textit{Verified: \textcolor{red}{INCORRECT} (FPF, $t^*=2$)}
\end{itemize}
\textbf{Verifier Guidance at $t^*=2$:}\\
\textit{``The factors of $+4$ summing to $-5$ must both be negative: $(-4)\times(-1)=4$ and $(-4)+(-1)=-5$. Using $+1$ is a sign error; the correct factoring is $(u-4)(u-1)=0$.''}
\end{tcolorbox}

\begin{tcolorbox}[title={Path A: Rectification from $t^*{=}2$ (same strategy corrected)}, fonttitle=\bfseries, colback=green!2!white, colframe=green!45!black, arc=2mm, boxrule=0.6pt]
\small
\begin{itemize}
  \item \texttt{<step 2'>} $(u-4)(u-1)=0 \Rightarrow u=4$ or $u=1$.
  \item \texttt{<step 3>} $2^x=4 \Rightarrow x=2$;\enspace $2^x=1 \Rightarrow x=0$.
\end{itemize}
\textbf{Final Answer:} $\boxed{x\in\{0,2\}}$ \hfill \textit{Status: \textcolor{green}{SUCCESS}}\\
\textit{Strategy: substitution with corrected factoring.}
\end{tcolorbox}

\begin{tcolorbox}[title={Path B: Exploration from $t{=}1$ (structurally different strategy)}, fonttitle=\bfseries, colback=blue!2!white, colframe=blue!40!black, arc=2mm, boxrule=0.6pt]
\small
\begin{itemize}
  \item \texttt{<step 1'>} Recognise $2^{2x}=(2^x)^2$. Factor the original expression directly as a quadratic in $2^x$: $(2^x-4)(2^x-1)=0$.
  \item \texttt{<step 2'>} $2^x=4 \Rightarrow x=2$;\enspace $2^x=1 \Rightarrow x=0$.
\end{itemize}
\textbf{Final Answer:} $\boxed{x\in\{0,2\}}$ \hfill \textit{Status: \textcolor{green}{SUCCESS}}\\
\textit{Strategy: direct factoring without substitution --- a structurally different path.}
\end{tcolorbox}

Path~A corrects the arithmetic error within the substitution framework; Path~B bypasses substitution entirely by recognising the product structure of the original expression.
Both paths reach the same correct answer via qualitatively different algebraic strategies, validating that multi-point redirection broadens the generator's solution repertoire rather than merely patching individual mistakes.

\subsection{Case Study III: Latent Drift and Recovery from Pre-Failure Anchors}
Errors in multi-step reasoning are often not purely local: a step that \emph{appears} correct may reflect a suboptimal earlier choice whose consequences only manifest steps later, a phenomenon we term \emph{latent drift}.
Single-point redirection from the FPF corrects the visible symptom but leaves the underlying fragility in place.
Multi-Point Redirection, by also redirecting from pre-FPF anchors, allows the generator to discover cleaner paths that avoid the root cause entirely.

\begin{tcolorbox}[title=Initial Attempt (Failed), fonttitle=\bfseries, colback=gray!3!white, colframe=gray!55!black, arc=2mm, boxrule=0.6pt]
\small
\textbf{Problem:} Solve for real $x$: $(x-1)^2 + (x+3)^2 = 20$.
\begin{itemize}
  \item \texttt{<step 1>} Expand: $(x^2-2x+1)+(x^2+6x+9)=20 \Rightarrow 2x^2+4x+10=20$. \hfill \textit{Verified: CORRECT}
  \item \texttt{<step 2>} Simplify: $2x^2+4x-10=0 \Rightarrow x^2+2x-5=0$. \hfill \textit{Verified: CORRECT}
  \item \texttt{<step 3>} Quadratic formula: $x=\frac{-2\pm\sqrt{4+20}}{2}=\frac{-2\pm\sqrt{24}}{2}$. Simplify $\sqrt{24}=2\sqrt{5}$, giving $x=-1\pm\sqrt{5}$. \hfill \textit{Verified: \textcolor{red}{INCORRECT} (FPF, $t^*=3$)}
\end{itemize}
\textbf{Verifier Guidance at $t^*=3$:}\\
\textit{``$\sqrt{24}=\sqrt{4\cdot 6}=2\sqrt{6}$, not $2\sqrt{5}$. The correct simplification gives $x=-1\pm\sqrt{6}$.''}
\end{tcolorbox}

\begin{tcolorbox}[title={Path A: Rectification from $t^*{=}3$ (FPF correction)}, fonttitle=\bfseries, colback=green!2!white, colframe=green!45!black, arc=2mm, boxrule=0.6pt]
\small
\begin{itemize}
  \item \texttt{<step 3'>} $x=\frac{-2\pm\sqrt{24}}{2}=\frac{-2\pm 2\sqrt{6}}{2}=-1\pm\sqrt{6}$.
\end{itemize}
\textbf{Final Answer:} $\boxed{x=-1\pm\sqrt{6}}$ \hfill \textit{Status: \textcolor{green}{SUCCESS}}\\
\textit{Corrects the radical simplification error, but retains the algebraically heavy expansion from Step~1.}
\end{tcolorbox}

\begin{tcolorbox}[title={Path B: Exploration from $t{=}1$ (avoiding the root cause)}, fonttitle=\bfseries, colback=blue!2!white, colframe=blue!40!black, arc=2mm, boxrule=0.6pt]
\small
\begin{itemize}
  \item \texttt{<step 1'>} Let $m=x+1$. Then $(x-1)=m-2$ and $(x+3)=m+2$, so $(m-2)^2+(m+2)^2=20$.
  \item \texttt{<step 2'>} Expand: $2m^2+8=20 \Rightarrow m^2=6 \Rightarrow m=\pm\sqrt{6}$.
  \item \texttt{<step 3'>} $x=m-1=-1\pm\sqrt{6}$.
\end{itemize}
\textbf{Final Answer:} $\boxed{x=-1\pm\sqrt{6}}$ \hfill \textit{Status: \textcolor{green}{SUCCESS}}\\
\textit{Exploits the symmetry of the expression via the substitution $m=x+1$, bypassing the discriminant computation entirely.}
\end{tcolorbox}

Path~A repairs the arithmetic error at the detected failure step.
However, the brute-force expansion introduced in Step~1 remains a latent source of fragility: similar problems would again require heavy discriminant arithmetic.
Path~B, redirected from $t=1$, introduces a symmetry-aware substitution $m=x+1$ that recognises the geometric structure of the sum of squares, rendering the subsequent arithmetic trivial and eliminating the error-prone radical simplification altogether.
The root cause of the failure was not the wrong simplification at Step~3, but the choice of brute-force expansion at Step~1 that created a more error-prone algebraic path.
This case demonstrates that multi-point redirection does not only fix local errors: by exploring from pre-FPF anchors, it uncovers more robust reasoning strategies that single-point redirection, restricted to the FPF alone, cannot reach.



\end{document}